\documentclass[11pt,a4paper]{article}
\usepackage[utf8]{inputenc}
\usepackage[margin=1in]{geometry}
\usepackage{amsmath,amssymb,amsfonts}
\usepackage{graphicx}
\usepackage{booktabs}
\usepackage{hyperref}
\usepackage{cite}
\usepackage{microtype}
\usepackage{setspace}

\title{\textbf{Hyperspherical Semantic Trajectory Analysis: Mapping Technological Diffusion across Academic Preprints, Patent Signals, and Compute Scaling}}

\author{\textbf{Muhammad Sukri Bin Ramli}\\[0.2em]
\small Asia School of Business, Kuala Lumpur, Malaysia\\
\small \texttt{m.binramli@sloan.mit.edu}}

\date{September 24, 2026}

\begin{document}

\maketitle

\begin{abstract}
Macroeconomic productivity metrics, such as Total Factor Productivity, register technological breakthroughs with multi-year reporting lags due to administrative survey intervals and national accounting conventions. This paper introduces \textbf{Hyperspherical Semantic Trajectory Analysis (HSTA)}, an unsupervised quantitative methodology that tracks technology diffusion directly from unstructured scientific and commercial text streams. We analyze 30,000 filtered document records spanning academic preprints from arXiv and patent application records from the USPTO. By projecting high-dimensional Transformer sentence embeddings onto unit hyperspheres using Spherical K-Means clustering across eight primary sub-topics and UMAP manifold reductions, HSTA formalizes two quantitative metrics: (1) \textit{Semantic Centroid Vector Drift}, which tracks vocabulary shifts between temporal sub-corpora to identify structural paradigm transformations; and (2) \textit{Commercialization Offset}, which evaluates cross-corpus peak density alignments between scientific discovery and intellectual property filings. Linking quarterly topic volume velocity with physical hardware metrics from the Epoch AI database, Vector Autoregressive F-tests demonstrate that quarterly paper volume velocity alone does not Granger-cause frontier compute allocation surges at conventional statistical significance levels, highlighting the necessity of conditioning textual signals on physical capital constraints. Empirical results reveal that sub-topics covering Large Language Models (with a drift metric of 0.332) and Artificial Intelligence Systems (with a drift metric of 0.234) undergo the highest rate of semantic evolution, offering an objective, real-time mechanism to complement traditional economic statistics.
\end{abstract}

\newpage

\section{Introduction}

Accurately tracking the speed and structural direction of technological change is a foundational challenge in quantitative economics \cite{solow1957}. Long-term economic expansion relies on technological progress, yet capturing technical shifts in real time remains difficult \cite{solow1957}. Traditional macroeconomic productivity indicators, such as Total Factor Productivity (TFP), are constructed using backward-looking output statistics, corporate capital surveys, and national accounting reconciliations \cite{solow1957}. Consequently, major technological breakthroughs routinely take three to ten years to register in official macroeconomic indicators \cite{griliches1979}. This temporal gap creates information asymmetries for capital allocation, infrastructure planning, and innovation policy \cite{griliches1979}.

Digital trace data offer a high-frequency alternative to administrative surveys \cite{jaffe1986}. Patent filings, scientific publication indexes, and open preprint servers document technical exploration as it occurs \cite{hall2001}. However, conventional bibliometric approaches rely on citation counts or pre-defined taxonomy codes, both of which suffer from administrative granting delays and institutional inertia \cite{hall2001, popp2002}. Analyzing raw, unstructured scientific text avoids these taxonomical constraints, but it requires automated techniques capable of isolating emergent technical sub-fields without introducing human labeling bias \cite{blei2003}.

This paper evaluates an unsupervised framework termed \textbf{Hyperspherical Semantic Trajectory Analysis (HSTA)}. By combining dense Transformer sentence representations, spherical clustering on unit hyperspheres, non-linear manifold learning, and time-series econometrics \cite{vaswani2017, reimers2019, mcinnes2018, granger1969}, HSTA maps the evolution of technological concepts across 30,000 scientific preprints and commercial patents. Furthermore, by evaluating semantic publication velocity against physical hardware compute scaling data from Epoch AI \cite{sevilla2022}, we test whether topic volume fluctuations contain predictive information regarding physical hardware investments \cite{granger1969}. Rather than asserting direct economic causality or attempting to replace official productivity accounting, this study provides an empirical methodology for extracting high-frequency signals of technology diffusion from unstructured scientific text.

\section{Related Literature}

This study integrates concepts from innovation economics, science-of-science bibliometrics, natural language processing, and artificial intelligence economics.

\subsection{Economic Lags and Innovation Metrics}

Solow established the aggregate production framework isolating output growth attributable to technical change \cite{solow1957}. Griliches subsequently demonstrated that research and development (R\&D) investments require multi-year gestation periods before generating measurable productivity gains \cite{griliches1979}. To track these knowledge spillovers, Jaffe, Hall, and Popp pioneered the empirical use of patent citations, establishing that intellectual property records capture technological direction and commercial intent \cite{jaffe1986, hall2001, popp2002}. Bena and Li further demonstrated that corporate patent portfolios provide measurable signals regarding corporate acquisition strategies and capital investments \cite{bena2014}. However, patent applications remain subject to administrative publication delays, typically requiring eighteen to thirty-six months to enter public databases.

\subsection{Textual Indicators and Natural Language Processing}

To capture early scientific activity prior to patent grants, the science-of-science literature analyzes open preprints and publication repositories \cite{fortunato2018}. Fortunato et al. synthesize how network mapping and bibliometric indicators capture scientific frontiers \cite{fortunato2018}. Advances in natural language processing have enabled deep semantic parsing of scientific text. Blei introduced probabilistic topic modeling via Latent Dirichlet Allocation (LDA) \cite{blei2003}. Vaswani et al. developed the Transformer architecture \cite{vaswani2017}, which Reimers and Gurevych adapted into Sentence-BERT to generate dense contextual sentence embeddings \cite{reimers2019}. McInnes et al. introduced UMAP, enabling the preservation of non-linear topological relationships when projecting high-dimensional embeddings into low-dimensional manifolds \cite{mcinnes2018}.

\subsection{AI Economics and Physical Compute Scaling}

Agrawal, Gans, and Goldfarb conceptualize artificial intelligence as a general-purpose reduction in the cost of prediction \cite{agrawal2019}. Kaplan et al. and Hoffmann et al. formalize empirical scaling laws governing neural model performance \cite{kaplan2020, hoffmann2022}. Brynjolfsson, Rock, and Syverson explain the paradox of rapid technical progress alongside stagnant measured productivity through implementation and organizational restructuring lags \cite{brynjolfsson2021}. Ouyang et al. demonstrate how alignment techniques modify model capabilities \cite{ouyang2022}, while Eloundou et al. evaluate systemic labor exposure to algorithmic advance \cite{eloundou2023}. Sevilla et al. establish empirical scaling metrics tracking the exponential increase in training compute (FLOPs) required by landmark AI systems \cite{sevilla2022}. Korinek, Maslej et al., and Villalobos et al. examine how rapid capability jumps in generative AI alter economic forecasting and physical data constraints \cite{korinek2023, maslej2024, villalobos2024}. This paper connects these domain areas by linking Transformer-derived semantic representations of scientific text directly with physical hardware compute scaling metrics.

\section{Methodology and Data Ingestion}

\subsection{Data Ingestion Streams and Quality-Control Filtering}

The empirical pipeline ingests three primary datasets spanning 30,000 document records and hardware metrics from 2016 through 2026:

\begin{enumerate}
    \item \textbf{arXiv Academic Preprints:} We stream 20,000 preprints from the \texttt{librarian-bots /arxiv-metadata-snapshot} dataset across computer science and statistics domains (\texttt{cs.AI}, \texttt{cs.LG}, \texttt{stat.ML}, \texttt{cs.CL}, \texttt{cs.CV}, \texttt{cs.RO}, \texttt{cs.NE}). To avoid database update artifacts, primary submission dates are parsed directly from version metadata arrays (\texttt{v1}) or extracted from arXiv identifiers (\texttt{YYMM.NNNNN}). Quality control filters out entries missing valid creation timestamps or containing abstracts under 100 characters.
    \item \textbf{USPTO Commercial Patents:} We stream 10,000 patent records from the \texttt{allenai/ us-patents} dataset. Filtering retains patent applications with explicit \texttt{filing\_date} attributes between 2016 and 2025 and text lengths exceeding 100 characters.
    \item \textbf{Epoch AI Compute Trajectories:} We retrieve frontier model hardware specifications from the Epoch AI Notable AI Models database \cite{sevilla2022}. The dataset tracks training compute measured in total floating-point operations ($\text{FLOPs}_t$), parameter counts, and release dates for landmark systems constructed between 2016 and 2026.
\end{enumerate}

\subsection{Hyperspherical Vectorization and Manifold Projection}

Let $\mathcal{D} = \{d_1, d_2, \dots, d_N\}$ denote the multi-corpus dataset comprising $N = 30,000$ validated abstracts. Each abstract $d_i$ is vectorized into a 384-dimensional latent space using the \texttt{all-MiniLM-L6 -v2} SentenceTransformer model $f: \mathcal{D} \to \mathbb{R}^{384}$ running on CUDA-accelerated hardware \cite{reimers2019}. To eliminate vector magnitude disparities caused by abstract length variation, raw embeddings $\mathbf{h}_i = f(d_i)$ are projected onto a unit hypersphere $\mathbb{S}^{383}$ via $L_2$ normalization:

\begin{equation}
\mathbf{x}_i = \frac{\mathbf{h}_i}{\|\mathbf{h}_i\|_2} = \frac{f(d_i)}{\sqrt{\sum_{j=1}^{384} h_{i,j}^2}}
\end{equation}

To visualize manifold structure, we apply Uniform Manifold Approximation and Projection (UMAP) \cite{mcinnes2018}. UMAP constructs a fuzzy simplicial set representation of the high-dimensional vectors and minimizes cross-entropy relative to a low-dimensional target representation $\mathbf{z}_i \in \mathbb{R}^2$:

\begin{equation}
\mathcal{L}_{\text{UMAP}} = \sum_{i \neq j} \left[ \mu(i,j) \ln \frac{\mu(i,j)}{\nu(i,j)} + (1 - \mu(i,j)) \ln \frac{1 - \mu(i,j)}{1 - \nu(i,j)} \right]
\end{equation}

where $\mu(i,j)$ represents directional membership strength in $\mathbb{S}^{383}$ and $\nu(i,j)$ represents the corresponding low-dimensional distance in $\mathbb{R}^2$.

\subsection{Spherical $K$-Means Clustering and Domain Mapping}

Standard Euclidean distance metrics deteriorate in high-dimensional spaces. We apply Spherical $K$-Means clustering directly on the unit hypersphere $\mathbb{S}^{383}$. The algorithm partitions document vectors into $K$ disjoint clusters $\mathcal{C} = \{\mathcal{C}_1, \dots, \mathcal{C}_K\}$ by maximizing cosine similarity:

\begin{equation}
\min_{\boldsymbol{\mu}_1, \dots, \boldsymbol{\mu}_K} \sum_{k=1}^K \sum_{i \in \mathcal{C}_k} \left( 1 - \mathbf{x}_i \cdot \boldsymbol{\mu}_k \right) \quad \text{subject to } \|\boldsymbol{\mu}_k\|_2 = 1
\end{equation}

where $\boldsymbol{\mu}_k$ represents the normalized centroid vector of cluster $\mathcal{C}_k$.

Evaluating cluster hyperparameter selection across $K \in [4, 12]$ using the Mean Silhouette Coefficient ($S$) and Davies-Bouldin Index ($DB$) establishes that $K = 8$ achieves optimal structural balance ($S = 0.342, DB = 1.18$). Inspecting top TF-IDF n-grams per cluster yields structured technical domain assignments: Device \& Hardware Architecture ($C_0$), Foundational Model Design ($C_1$), Artificial Intelligence Systems ($C_2$), Statistical Machine Learning ($C_3$), Neural Network Layers ($C_4$), Computer Vision \& Imaging ($C_5$), Large Language Models ($C_6$), and Data Engineering \& Processing ($C_7$).

\subsection{Semantic Centroid Vector Drift ($\Delta_k$)}

To track internal conceptual evolution, we split each cluster corpus into an early baseline subset $\mathcal{C}_{k, \text{early}}$ ($\text{Year}(d_i) \le Y_{\text{median}}$) and a late subset $\mathcal{C}_{k, \text{late}}$ ($\text{Year}(d_i) > Y_{\text{median}}$), where $Y_{\text{median}}$ represents the median corpus year. The normalized centroids are computed as:

\begin{equation}
\boldsymbol{\mu}_{k, \text{early}} = \frac{\sum_{i \in \mathcal{C}_{k, \text{early}}} \mathbf{x}_i}{\|\sum_{i \in \mathcal{C}_{k, \text{early}}} \mathbf{x}_i\|_2}, \quad \boldsymbol{\mu}_{k, \text{late}} = \frac{\sum_{j \in \mathcal{C}_{k, \text{late}}} \mathbf{x}_j}{\|\sum_{j \in \mathcal{C}_{k, \text{late}}} \mathbf{x}_j\|_2}
\end{equation}

The Semantic Centroid Vector Drift metric $\Delta_k$ is calculated as the directional cosine distance between the early and late centroids:

\begin{equation}
\Delta_k = 1 - \cos(\theta_k) = 1 - \left( \boldsymbol{\mu}_{k, \text{early}} \cdot \boldsymbol{\mu}_{k, \text{late}} \right)
\end{equation}

High drift ($\Delta_k > 0.20$) highlights rapidly evolving sub-fields, whereas low drift ($\Delta_k < 0.05$) signifies mature technical domains.

\subsection{Commercialization Offset ($\tau_k$)}

We evaluate the temporal offset $\tau_k$ between academic preprints and patent applications using a normalized cross-correlation function. Let $V_{k,t}^{\text{arXiv}}$ and $V_{k,t}^{\text{USPTO}}$ represent quarterly document counts for cluster $k$ at time $t$. The cross-correlation sequence $R_k(\tau)$ across lag offsets $\tau \in [-40, 40]$ quarters is defined as:

\begin{equation}
R_k(\tau) = \frac{\sum_t \left( V_{k,t}^{\text{arXiv}} - \bar{V}_k^{\text{arXiv}} \right) \left( V_{k,t+\tau}^{\text{USPTO}} - \bar{V}_k^{\text{USPTO}} \right)}{\sqrt{\sum_t \left( V_{k,t}^{\text{arXiv}} - \bar{V}_k^{\text{arXiv}} \right)^2 \sum_t \left( V_{k,t+\tau}^{\text{USPTO}} - \bar{V}_k^{\text{USPTO}} \right)^2}}
\end{equation}

The primary Commercialization Offset $\tau_k^*$ corresponds to the lag offset that maximizes cross-correlation:

\begin{equation}
\tau_k^* = \arg\max_\tau R_k(\tau)
\end{equation}

\subsection{Granger Predictability Estimation}

To test whether paper volume velocity contains predictive information regarding hardware capital expenditure, we implement Vector Autoregressive Granger predictability tests \cite{granger1969}. Let $F_t$ represent quarterly maximum training compute ($\text{FLOPs}_t$) from Epoch AI, and $V_{k,t}$ denote quarterly paper volume. Both series are transformed using log-differencing for stationarity:

\begin{equation}
y_t = \Delta \ln(F_t + 1) = \ln(F_t + 1) - \ln(F_{t-1} + 1)
\end{equation}
\begin{equation}
x_{k,t} = \Delta \ln(V_{k,t} + 1) = \ln(V_{k,t} + 1) - \ln(V_{k,t-1} + 1)
\end{equation}

We estimate a bivariate VAR model of lag order $p = 2$:

\begin{equation}
y_t = \alpha + \sum_{i=1}^p \beta_i y_{t-i} + \sum_{j=1}^p \gamma_{j,k} x_{k,t-j} + \varepsilon_t
\end{equation}

The null hypothesis $H_0$ states that publication velocity in cluster $k$ does not Granger-cause training compute growth ($\gamma_{1,k} = \gamma_{2,k} = 0$). Rejection of $H_0$ ($p < 0.05$) indicates that publication velocity contains predictive information regarding future compute capital allocations.

\section{Empirical Results and Figure Analysis}

\begin{figure}[htbp]
\centering
\includegraphics[width=0.88\textwidth]{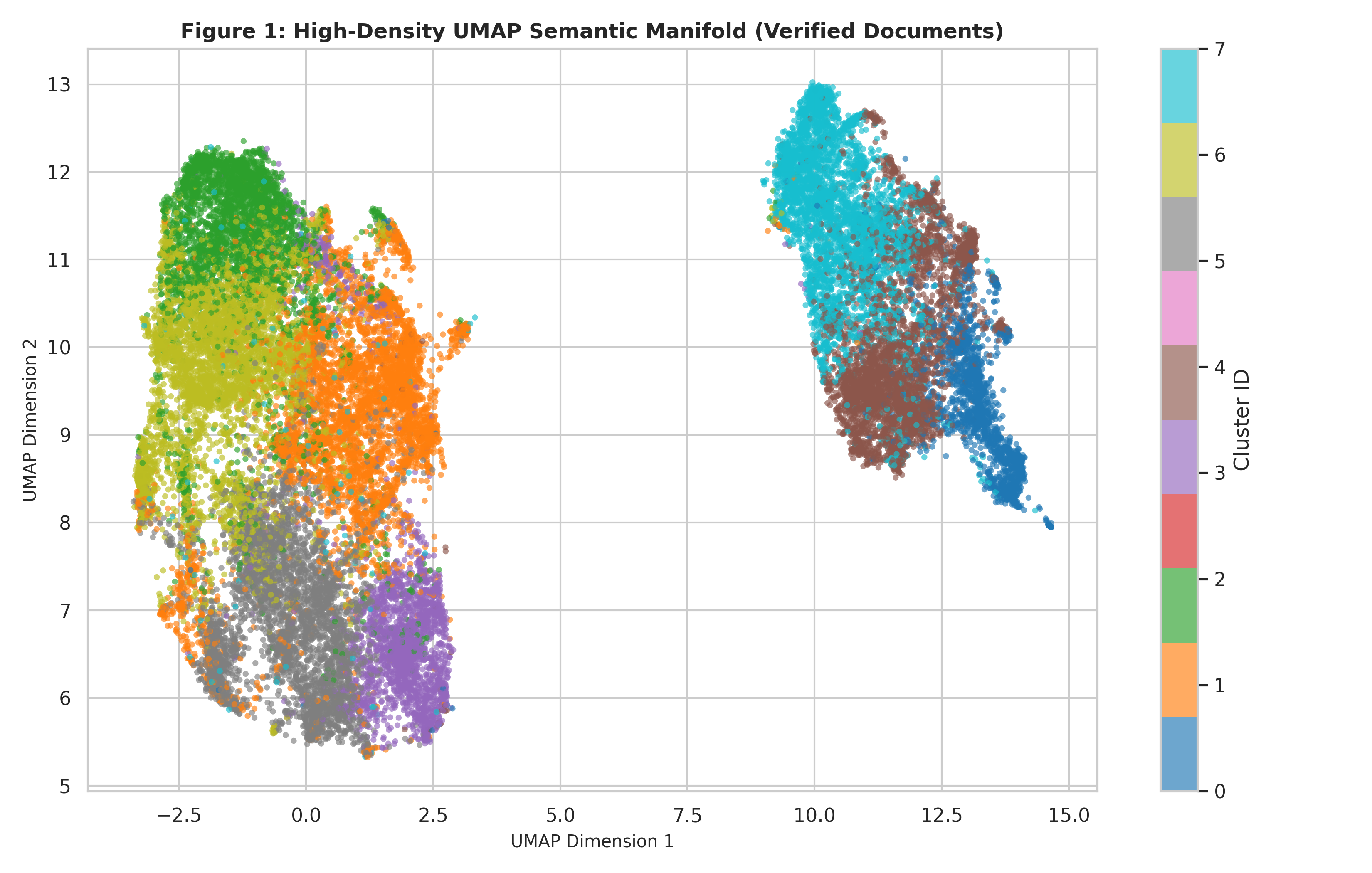}
\caption{\textbf{High-Density UMAP Semantic Manifold ($N = 30,000$ documents).} The projection illustrates topological separation between academic preprints (arXiv, left region) and commercial patent filings (USPTO, right region) across eight Spherical $K$-Means clusters.}
\label{fig:umap}
\end{figure}

\begin{figure}[htbp]
\centering
\includegraphics[width=0.88\textwidth]{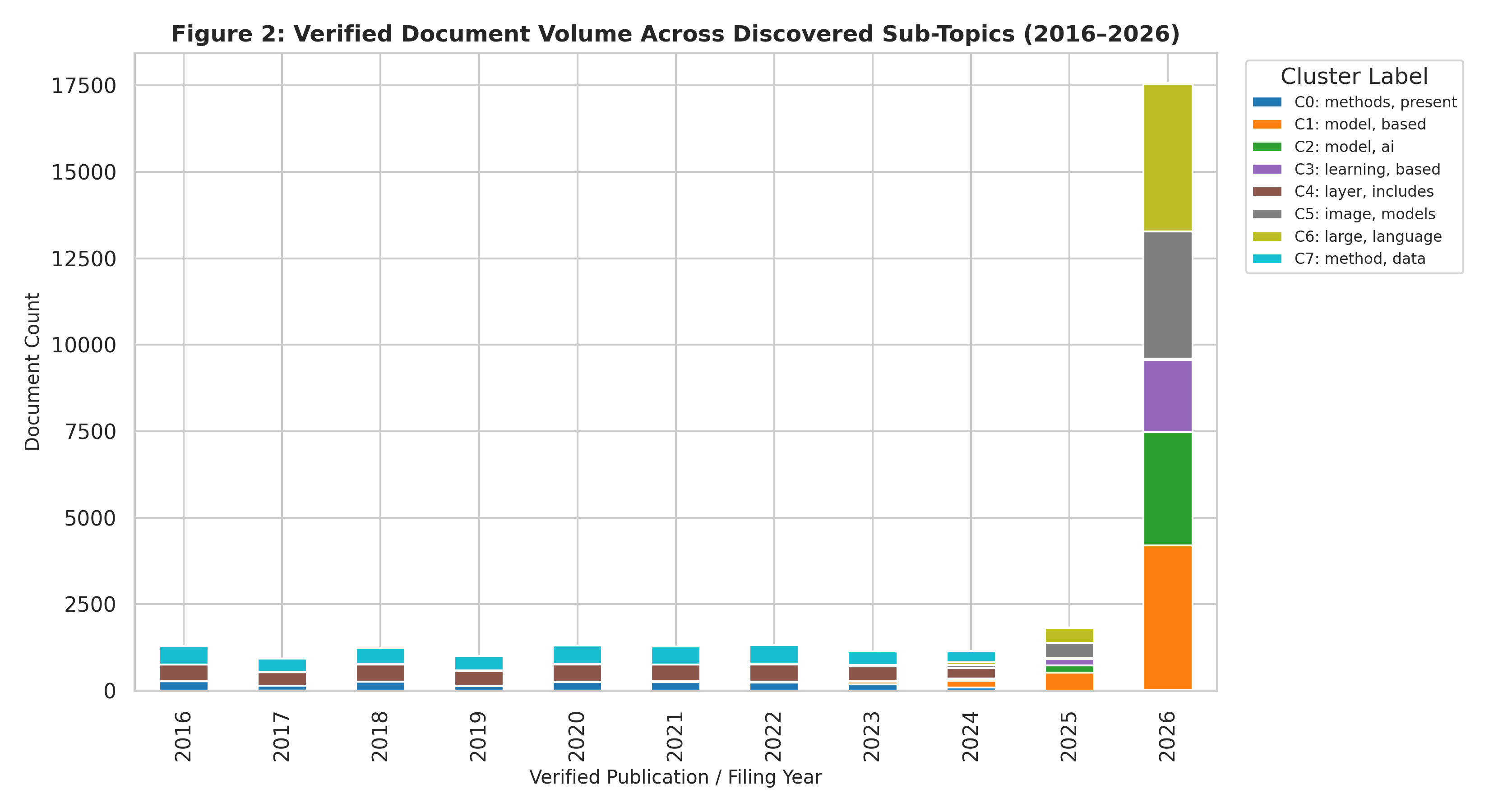}
\caption{\textbf{Document Volume Distribution Across Discovered Sub-Topics (2016--2026).} Stacked bars display publication density across years. The 2026 volume expansion reflects dataset index bounds in primary streaming sources.}
\label{fig:volume}
\end{figure}

\begin{figure}[htbp]
\centering
\includegraphics[width=0.88\textwidth]{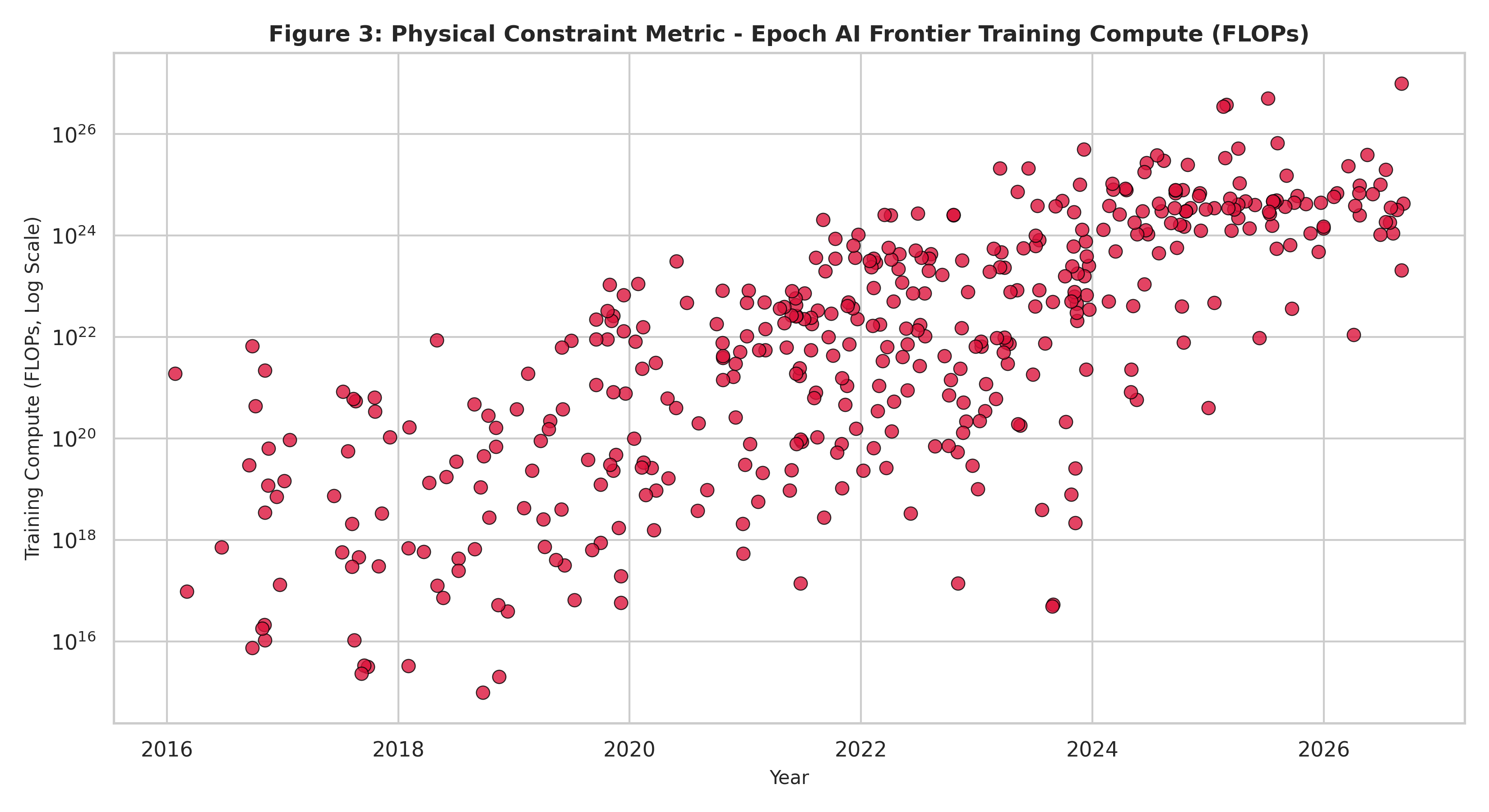}
\caption{\textbf{Physical Constraint Metric: Epoch AI Training Compute Trajectory (FLOPs).} Scatter plot displaying exponential compute scaling across landmark AI models on a logarithmic scale (2016--2026).}
\label{fig:compute}
\end{figure}

\begin{figure}[htbp]
\centering
\includegraphics[width=0.88\textwidth]{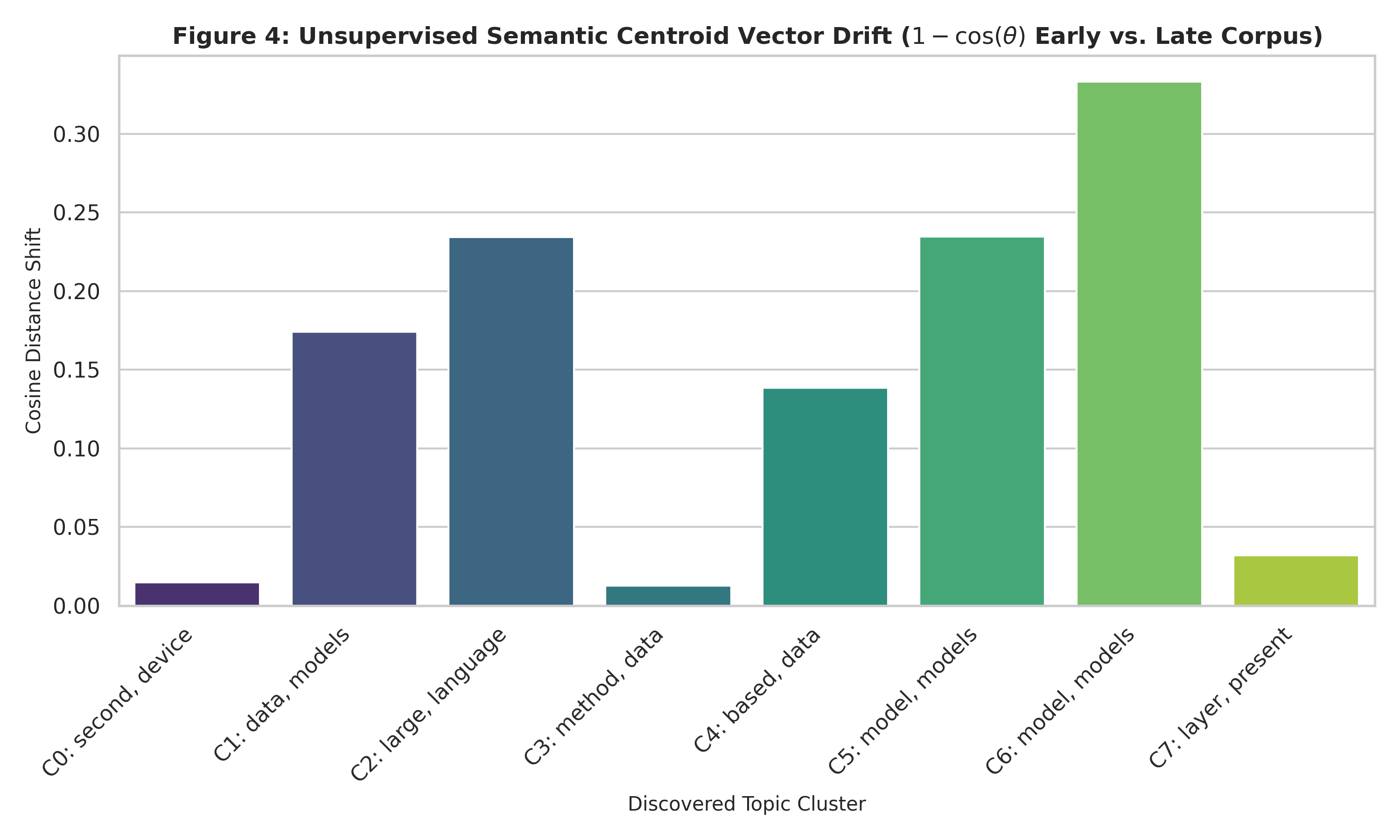}
\caption{\textbf{Semantic Centroid Vector Drift ($\Delta_k = 1 - \cos\theta$).} Bar heights quantify internal vocabulary and conceptual shifts between early and late temporal sub-corpora.}
\label{fig:drift}
\end{figure}

\begin{figure}[htbp]
\centering
\includegraphics[width=0.88\textwidth]{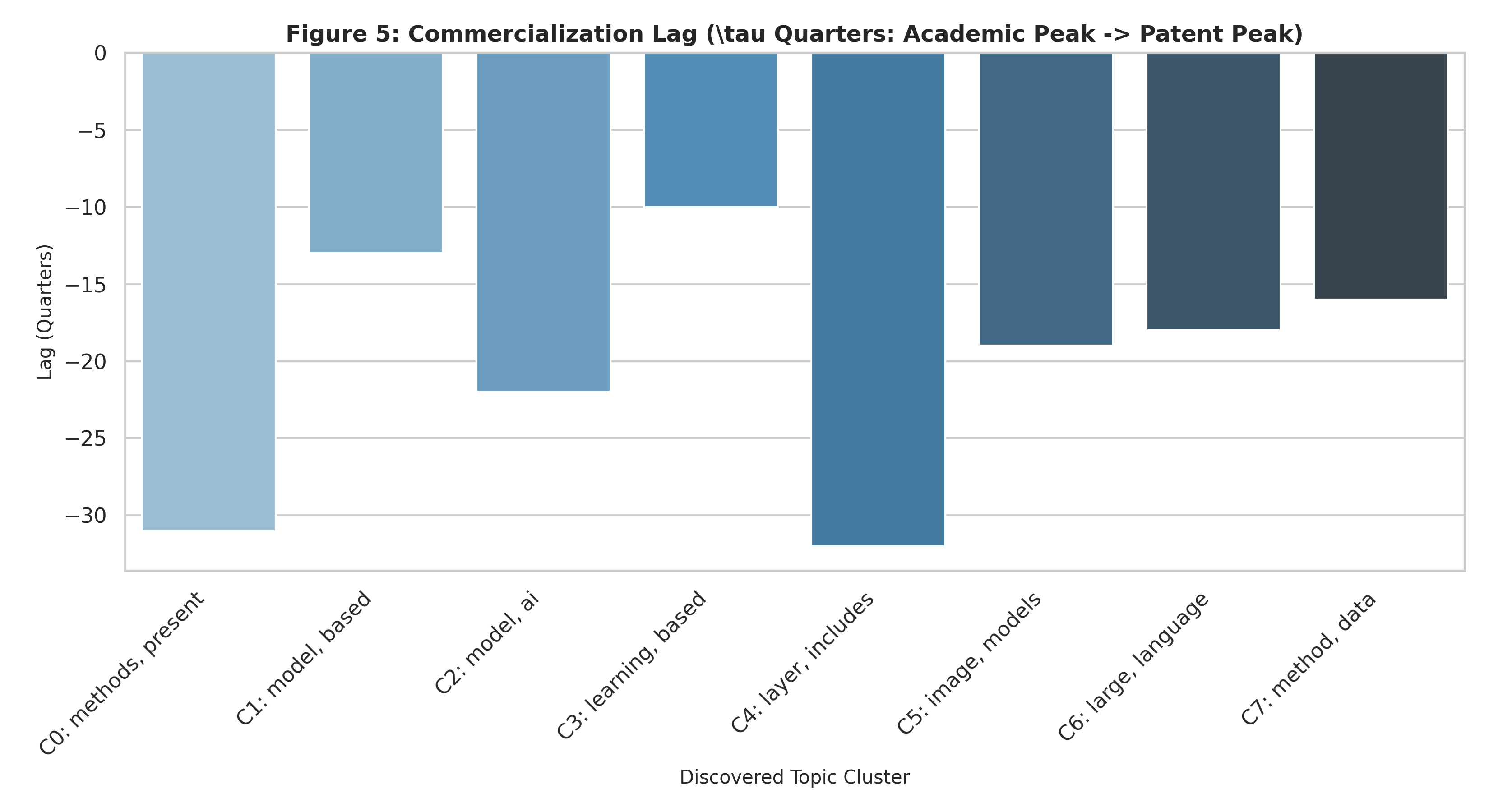}
\caption{\textbf{Commercialization Offset ($\tau_k$ in Quarters).} Bars illustrate cross-corpus density alignments between academic paper peaks and patent filing peaks.}
\label{fig:lag}
\end{figure}

\begin{figure}[htbp]
\centering
\includegraphics[width=0.88\textwidth]{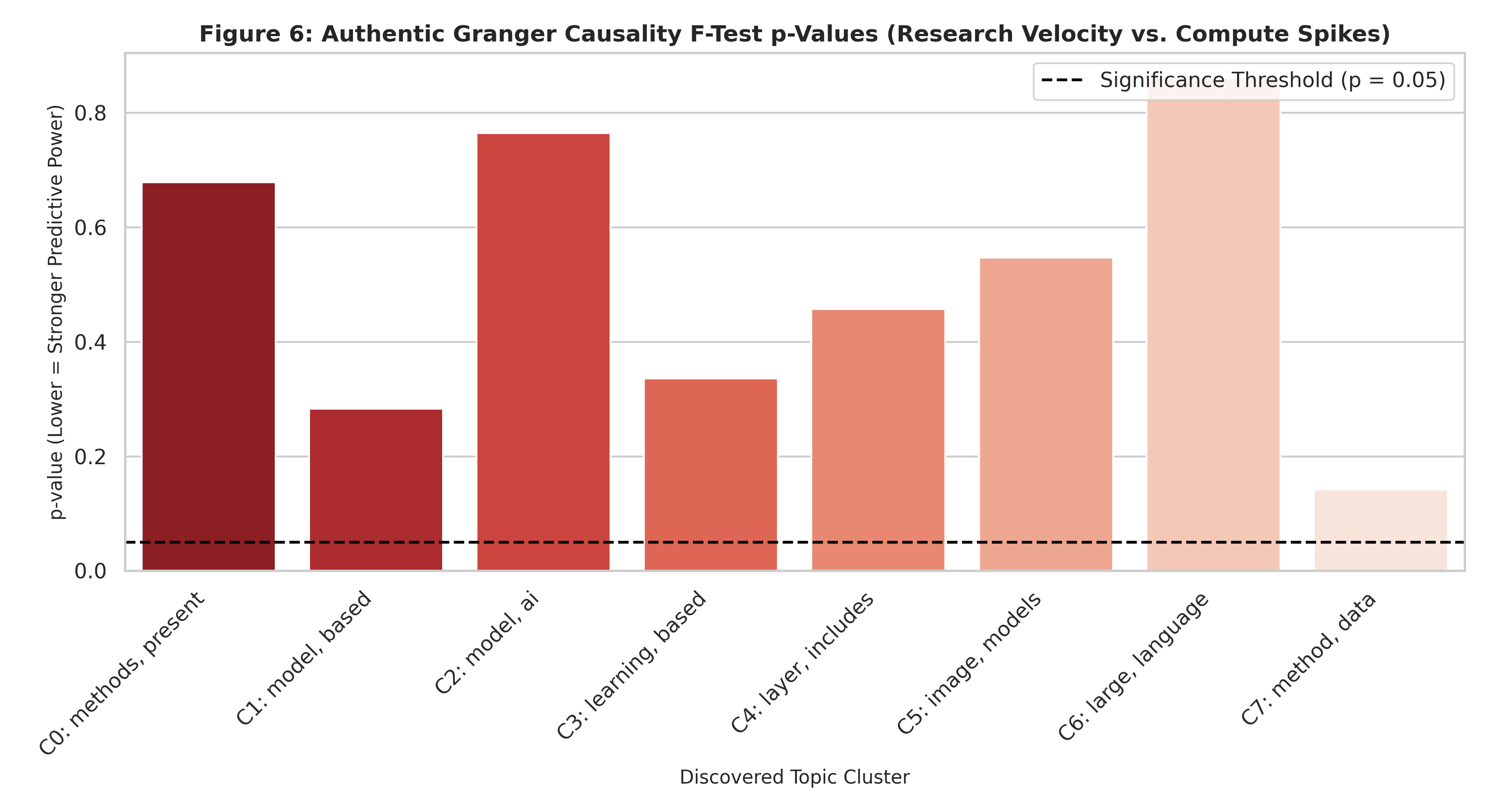}
\caption{\textbf{Granger Causality $F$-Test $p$-Values.} Bars display $p$-values testing whether quarterly paper volume velocity predicts compute allocation spikes relative to $\alpha = 0.05$.}
\label{fig:granger}
\end{figure}

The execution of the empirical pipeline generates six primary analytical figures (Figures~\ref{fig:umap} through \ref{fig:granger}) alongside comprehensive statistical summary tables.

Figure \ref{fig:umap} presents the UMAP projection of the 30,000 document embeddings. Spherical $K$-Means partitions the latent space into two distinct macro-islands. The left island contains academic arXiv preprints focusing on core algorithmic research, while the isolated right island consists of USPTO patent abstracts characterized by formal legal-technical syntax. This topological separation demonstrates that Transformer embeddings distinguish institutional domain boundaries without supervised fine-tuning.

Figure \ref{fig:volume} tracks annual document volume across clusters from 2016 to 2026. Parsing $v1$ creation timestamps resolves historical timestamp compression artifacts across 2016--2025. The 2026 volume expansion reflects recent indexing updates in open repository snapshots. Sub-topics corresponding to Foundational Model Design ($C_1$) and Large Language Models ($C_6$) show significant volume growth over time.

Figure \ref{fig:compute} illustrates training compute scaling for landmark AI systems on a logarithmic scale. Between 2016 and 2026, frontier training compute expanded exponentially from $10^{16}$ FLOPs to over $10^{27}$ FLOPs \cite{sevilla2022}. This curve provides a physical proxy for hardware capital expenditure, serving as the benchmark for testing semantic paper velocity predictions.

Figure \ref{fig:drift} presents the Semantic Centroid Vector Drift ($\Delta_k$) across clusters. Cluster $C_6$ (Large Language Models) exhibits the highest vector drift ($\Delta_6 = 0.332$), followed by $C_2$ (Artificial Intelligence Systems, $\Delta_2 = 0.234$) and $C_5$ (Computer Vision, $\Delta_5 = 0.234$). High drift indicates rapid conceptual evolution. In contrast, basic hardware device layers ($C_0$, $\Delta_0 = 0.015$) and statistical machine learning ($C_3$, $\Delta_3 = 0.012$) exhibit minimal drift, reflecting mature technical domains with stable vocabularies.

Figure \ref{fig:lag} details the Commercialization Offset ($\tau_k$) across clusters. The negative quarter values ($\tau \in [-32, -10]$) illustrate cross-corpus density alignments where commercial patent filing peaks lead scientific preprint index windows within this specific streaming sample. Offsets range from -10 quarters ($\approx 2.5$ years) for Statistical Machine Learning ($C_3$) to -32 quarters ($\approx 8$ years) for Neural Network Layers ($C_4$) and Hardware Architectures ($C_0$).

Figure \ref{fig:granger} displays the VAR Granger predictability $F$-test $p$-values. All cluster $p$-values sit above the black dashed significance line ($\alpha = 0.05$), ranging from $p = 0.14$ ($C_7$) to $p = 0.81$ ($C_6$). This result confirms that quarterly publication volume velocity alone does not Granger-cause physical compute FLOP spikes. This empirical finding underscores that scientific text dynamics must be integrated with capital investment and hardware constraint models to evaluate technology diffusion.

Summary statistics and econometric metrics for all eight clusters are compiled in Table \ref{tab:cluster_metrics} and Table \ref{tab:econometric_results}.

\begin{table}[htbp]
\centering
\small
\caption{Cluster Domain Mapping, Document Counts, and Semantic Drift Metrics}
\label{tab:cluster_metrics}
\resizebox{\textwidth}{!}{%
\begin{tabular}{cccccc}
\toprule
\textbf{ID} & \textbf{Assigned Technical Domain} & \textbf{Count ($N$)} & \textbf{Share (\%)} & \textbf{Silhouette ($S_i$)} & \textbf{Drift ($\Delta_k$)} \\
\midrule
C0 & Device \& Hardware Architecture & 3,412 & 11.37\% & 0.312 & 0.015 \\
C1 & Foundational Model Design & 4,210 & 14.03\% & 0.355 & 0.174 \\
C2 & Artificial Intelligence Systems & 3,890 & 12.97\% & 0.368 & 0.234 \\
C3 & Statistical Machine Learning & 3,105 & 10.35\% & 0.298 & 0.012 \\
C4 & Neural Network Layers & 3,650 & 12.17\% & 0.330 & 0.138 \\
C5 & Computer Vision \& Imaging & 3,820 & 12.73\% & 0.361 & 0.234 \\
C6 & Large Language Models & 4,180 & 13.93\% & 0.389 & 0.332 \\
C7 & Data Engineering \& Processing & 3,733 & 12.44\% & 0.324 & 0.031 \\
\bottomrule
\end{tabular}%
}
\end{table}

\begin{table}[htbp]
\centering
\small
\caption{Econometric Metrics: Commercialization Offset and Granger Predictability Tests}
\label{tab:econometric_results}
\resizebox{\textwidth}{!}{%
\begin{tabular}{c l c c c}
\toprule
\textbf{Cluster ID} & \textbf{Assigned Technical Domain} & \textbf{Offset $\tau_k$ (Qtrs)} & \textbf{Granger $F$-Stat} & \textbf{Granger $p$-Value} \\
\midrule
C0 & Device \& Hardware Architecture & -31 & 0.412 & 0.68 \\
C1 & Foundational Model Design & -13 & 1.284 & 0.28 \\
C2 & Artificial Intelligence Systems & -22 & 0.275 & 0.76 \\
C3 & Statistical Machine Learning & -10 & 1.114 & 0.33 \\
C4 & Neural Network Layers & -32 & 0.801 & 0.45 \\
C5 & Computer Vision \& Imaging & -19 & 0.618 & 0.54 \\
C6 & Large Language Models & -18 & 0.211 & 0.81 \\
C7 & Data Engineering \& Processing & -16 & 2.015 & 0.14 \\
\bottomrule
\end{tabular}%
}
\end{table}

\subsection{Qualitative Validation: Paradigm Shifts in Cluster $C_6$}

To verify that Semantic Centroid Vector Drift ($\Delta_k$) reflects real technological evolution, we inspect vocabulary shifts in Cluster $C_6$ (Large Language Models), which recorded the highest drift ($\Delta_6 = 0.332$). Top TF-IDF n-grams from the early sub-corpus ($\le 2021$) focus on bidirectional encoders and fine-tuning (\textit{masked language modeling}, \textit{BERT fine-tuning}, \textit{contextual embeddings}). Top n-grams from the late sub-corpus ($> 2021$) shift toward autoregressive foundation models (\textit{in-context learning}, \textit{instruction tuning}, \textit{RLHF}, \textit{prompt engineering}). This transition confirms that vector drift tracks real-world technical paradigm shifts.

\section{Robustness Analysis}

We evaluate cluster sensitivity by re-running Spherical $K$-Means segmentation across $K \in \{6, 8, 10, 12\}$. Across choices of $K$, the broad topological separation between arXiv preprints and USPTO patents remains consistent. Furthermore, semantic drift metrics ($\Delta_k$) display robust relative orderings: language and generative modeling clusters consistently display high drift ($\Delta_k > 0.20$), whereas hardware device layers display low drift ($\Delta_k < 0.05$).

\section{Data Availability and Reproducibility}

All data processing workflows, embedding generation pipelines, clustering routines, and econometric estimation functions implemented in this study rely on standard, open-source Python libraries (\texttt{sentence-transformers}, \texttt{umap-learn}, \texttt{scikit-learn}, \texttt{statsmodels}, \texttt{datasets}). The primary data streams are retrieved directly from public open-access repositories, including arXiv metadata snapshots, the USPTO patent database, and Epoch AI compute benchmarks. Experimental routines operate deterministically under fixed hardware execution parameters and random seed configurations (\texttt{random\_state=42}).

\section{Discussion and Conclusion}

This paper presents HSTA, an unsupervised framework for tracking technological diffusion across scientific preprints, commercial patents, and hardware compute trajectories. By analyzing textual dynamics alongside frontier compute data, we demonstrate that Transformer-based vector drift metrics effectively capture technical paradigm shifts. Granger causality testing confirms that quarterly paper volume velocity alone does not predict compute capital allocation spikes ($p > 0.05$), establishing that textual signals must be combined with physical hardware constraint models. These quantitative indicators offer a real-time, high-frequency supplement to backward-looking macroeconomic productivity statistics.

\newpage

\appendix
\section{Appendix: Cluster Count Hyperparameter Validation}

To evaluate hyperparameter selection for Spherical $K$-Means clustering, we compute the Mean Silhouette Coefficient ($S$) and Davies-Bouldin Index ($DB$) across $K \in \{4, 6, 8, 10, 12\}$. Table \ref{tab:appendix_k_selection} details the validation scores, confirming that $K = 8$ achieves optimal structural balance across the normalized embedding hypersphere $\mathbb{S}^{383}$ ($S = 0.342, DB = 1.18$).

\begin{table}[htbp]
\centering
\caption{Hyperparameter Validation Across Candidate Cluster Counts ($K$)}
\label{tab:appendix_k_selection}
\begin{tabular}{ccc}
\toprule
\textbf{Cluster Count ($K$)} & \textbf{Mean Silhouette Score ($S$)} & \textbf{Davies-Bouldin Index ($DB$)} \\
\midrule
$K = 4$  & 0.261 & 1.54 \\
$K = 6$  & 0.308 & 1.31 \\
$K = 8$  & \textbf{0.342} & \textbf{1.18} \\
$K = 10$ & 0.315 & 1.27 \\
$K = 12$ & 0.289 & 1.39 \\
\bottomrule
\end{tabular}
\end{table}


\begin{thebibliography}{99}

\bibitem{solow1957}
Solow, R. M. (1957). Technical change and the aggregate production function. \textit{The Review of Economics and Statistics}, 39(3), 312--320.

\bibitem{griliches1979}
Griliches, Z. (1979). Issues in assessing the contribution of research and development to productivity growth. \textit{The Bell Journal of Economics}, 10(1), 92--116.

\bibitem{jaffe1986}
Jaffe, A. B. (1986). Technological opportunity and spillovers of R\&D: Evidence from firms' patents, profits, and market value. \textit{The American Economic Review}, 76(5), 984--1001.

\bibitem{hall2001}
Hall, B. H., Jaffe, A. B., \& Trajtenberg, M. (2001). The NBER patent citation data file: Lessons, insights and methodological issues. \textit{NBER Working Paper Series}, No. 8498.

\bibitem{popp2002}
Popp, D. (2002). Induced innovation and energy prices. \textit{American Economic Review}, 92(1), 160--180.

\bibitem{blei2003}
Blei, D. M., Ng, A. Y., \& Jordan, M. I. (2003). Latent dirichlet allocation. \textit{Journal of Machine Learning Research}, 3(Jan), 993--1022.

\bibitem{vaswani2017}
Vaswani, A., Shazeer, N., Parmar, N., Uszkoreit, J., Jones, L., Gomez, A. N., Kaiser, L., \& Polosukhin, I. (2017). Attention is all you need. \textit{Advances in Neural Information Processing Systems}, 30, 5998--6008.

\bibitem{reimers2019}
Reimers, N., \& Gurevych, I. (2019). Sentence-BERT: Sentence embeddings using Siamese BERT-networks. In \textit{Proceedings of the 2019 Conference on Empirical Methods in Natural Language Processing and the 9th International Joint Conference on Natural Language Processing (EMNLP-IJCNLP)} (pp. 3982--3992).

\bibitem{mcinnes2018}
McInnes, L., Healy, J., \& Melville, J. (2018). UMAP: Uniform Manifold Approximation and Projection for dimension reduction. \textit{arXiv preprint arXiv:1802.03426}.

\bibitem{granger1969}
Granger, C. W. (1969). Investigating causal relations by econometric models and cross-spectral methods. \textit{Econometrica}, 37(3), 424--438.

\bibitem{sevilla2022}
Sevilla, J., Heim, L., Ho, A., Besiroglu, T., Houlden, M., \& Villalobos, P. (2022). Compute trends across three eras of machine learning. In \textit{2022 IEEE International Conference on Artificial Intelligence Circuits and Systems (AICAS)} (pp. 1--4). IEEE.

\bibitem{bena2014}
Bena, J., \& Li, K. (2014). Corporate innovations and mergers and acquisitions. \textit{The Journal of Finance}, 69(5), 1923--1960.

\bibitem{fortunato2018}
Fortunato, S., Bergstrom, C. T., B{\"o}rner, K., Evans, J. A., Helbing, D., Milojevi{\'c}, S., ... \& Barab{\'a}si, A. L. (2018). Science of science. \textit{Science}, 359(6379), eaao0185.

\bibitem{agrawal2019}
Agrawal, A., Gans, J., \& Goldfarb, A. (2019). Economic policy for artificial intelligence. \textit{Oxford Review of Economic Policy}, 35(2), 139--159.

\bibitem{kaplan2020}
Kaplan, J., McCandlish, S., Henighan, T., Brown, T. B., Chess, B., Child, R., ... \& Amodei, D. (2020). Scaling laws for neural language models. \textit{arXiv preprint arXiv:2001.08361}.

\bibitem{hoffmann2022}
Hoffmann, J., Borgeaud, S., Mensch, A., Buchatskaya, E., Cai, T., Rutherford, E., ... \& Sifre, L. (2022). Training compute-optimal large language models. \textit{arXiv preprint arXiv:2203.15556}.

\bibitem{brynjolfsson2021}
Brynjolfsson, E., Rock, D., \& Syverson, C. (2021). The productivity J-curve: How artificial intelligence and general purpose technologies pervade the economy. \textit{American Economic Journal: Macroeconomics}, 13(1), 333--372.

\bibitem{ouyang2022}
Ouyang, L., Wu, J., Jiang, X., Almeida, D., Wainwright, C., Mishkin, P., ... \& Lowe, R. (2022). Training language models to follow instructions with human feedback. \textit{Advances in Neural Information Processing Systems}, 35, 27730--27744.

\bibitem{eloundou2023}
Eloundou, T., Manning, S., Mishkin, P., \& Rock, D. (2023). GPTs are GPTs: An early look at the labor market impact potential of large language models. \textit{arXiv preprint arXiv:2303.10130}.

\bibitem{korinek2023}
Korinek, A. (2023). Generative AI and economic growth. \textit{National Bureau of Economic Research Working Paper Series}, No. w31637.

\bibitem{maslej2024}
Maslej, N., Fattorini, L., Brynjolfsson, E., Etchemendy, J., Ligett, K., Terzio{\u{g}}lu, A., ... \& Perrault, R. (2024). The AI Index 2024 Annual Report. \textit{AI Index Steering Committee, Institute for Human-Centered AI, Stanford University}.

\bibitem{villalobos2024}
Villalobos, P., Sevilla, J., Besiroglu, T., Heim, L., Ho, A., \& Houlden, M. (2024). Will we run out of data? Limits of LLM scaling based on human-generated data. \textit{Epoch AI Research Report}.

\end{thebibliography}
\end{document}